\documentclass[onecolumn,10pt]{article}

\usepackage{graphicx}
\usepackage[utf8]{inputenc}
\usepackage{url}            
\usepackage{booktabs}       
\usepackage{amsfonts}       
\usepackage{nicefrac}       
\usepackage{microtype}      
\usepackage{amsmath,amssymb}
\usepackage{float}
\usepackage[USenglish]{babel}
\usepackage{dsfont}
\usepackage{caption}
\usepackage{subcaption}
\usepackage{paralist}
\usepackage{algorithmicx}
\usepackage[ruled]{algorithm}
\usepackage{algpseudocode}
\usepackage{mathptmx}
\usepackage{pict2e}
\usepackage{fullpage}
\usepackage{colorprofiles}
\usepackage{textcomp}
\usepackage{xcolor}
\usepackage{array}
\usepackage[caption=false,font=normalsize,labelfont=sf,textfont=sf]{subfig}
\usepackage{verbatim}
\usepackage{cite}
\usepackage{tabularx}
\usepackage{lmodern} 
\usepackage{times}
\usepackage{cite}
\usepackage{moreverb}
\usepackage{bm}
\usepackage{subfig}
\renewcommand\vec{\mathbf}

\newcommand{\hsb}{\hspace{.25in}}
\newcommand{\hsc}{\hspace{.40in}}
\newcommand{\hsd}{\hspace{.55in}}
\newcommand{\hse}{\hspace{.70in}}

\newcommand{\bbmp}{\begin{boxedminipage}}
\newcommand{\ebmp}{\end{boxedminipage}}
\newcommand{\bmp}{\begin{minipage}}
\newcommand{\emp}{\end{minipage}}

\renewcommand\vec{\mathbf}

\newcommand{\R}{\mathbb{R}}

\DeclareMathOperator{\argmax}{arg\,max} 

\def\BibTeX{{\rm B\kern-.05em{\sc i\kern-.025em b}\kern-.08em
    T\kern-.1667em\lower.7ex\hbox{E}\kern-.125emX}}

\newtheorem{definition}{Definition}

\newtheorem{lemma}{Lemma}

\newtheorem{corollary}{Corollary}
\newenvironment{proof}{\paragraph{Proof:}}{\hfill$\square$}
\title{Novel Deep Neural Network Classifier Characterization Metrics with Applications to Dataless Evaluation }
\author{
Nathaniel Dean and Dilip Sarkar\\
Computer Science \\
University of Miami\\
Coral Gables, FL 33124\\
Email: nxd551@miami.edu\\
Email: sarkar@miami.edu\\
}
\begin{document}
\maketitle

\begin{abstract}
The mainstream AI community has seen a rise in large-scale open-source classifiers, often pre-trained on vast datasets and tested on standard benchmarks; however, users facing diverse needs and limited, expensive test data may be overwhelmed by available choices. Deep Neural Network (DNN) classifiers undergo training, validation, and testing phases using example dataset, with the testing phase focused on determining the classification accuracy of test examples without delving into the inner working of the classifier. In this work, we evaluate a DNN classifier's training quality without any example dataset. It is assumed that a DNN is a composition of a feature extractor and a classifier which is the penultimate completely connected layer. The quality of a classifier is estimated using its weight vectors. The feature extractor is characterized using two metrics that utilize feature vectors it produces when synthetic data is fed as input. These synthetic input vectors are produced by backpropagating desired outputs of the classifier. Our empirical study of the proposed method for ResNet18, trained with CAFIR10 and CAFIR100 datasets, confirms that data-less evaluation of DNN classifiers is indeed possible.
\end{abstract}



\section{Introduction }
\label{sec:IntroductionAug23}
The mainstream artificial intelligence community has experienced a surge in the availability of large-scale open-source classifiers \cite{huggingface2024arxiv}, catering to diverse end-users with varying use cases.   Typically, these models are pre-trained on very large datasets and tested on some standard benchmark dataset, but end-users needing to select an open-source model may have a different use case and be overwhelmed by the possible choices.  Further, the end-user may only have a small amount of test data that may not accurately reflect their use-case or it would be costly to setup or bolster their data pipeline for every possible open-source model \cite{ActiveTesting2021Kossen,ActiveSurrogateEstimator2022Kossem}.  We propose a dataless evaluation method to pre-screen available models using only the weights and architecture of the network to eliminate poorly trained models and allow users to focus their data testing on only a few down-selected possibilities.
\par
Often Machine Learning (ML) models are trained, validated, and tested with example data. A significant part of the data is used to estimate the parameters of a model, next validation data is used for hyperparameter and model selection, and finally the test data is used to evaluate performance of the selected model \cite{ModelEvaluationModelSelection2020arXiv,EvaluatingMachineLearningModels2019eBook}. \emph{Deep Neural Networks} (DNNs) follow a similar method, where training examples are used to estimate parameters of a DNN. For a given example dataset, different model-architectures are  trained and using a validation dataset a model is selected. Modern DNN models have hyper-parameters and their good values are selected using validation datasets as well. Finally, performance of the selected DNN (network parameters and hyper-parameters) is tested using a test dataset.  
\par
Even when training, validation, and test datasets are coming from the same distribution or created by partitioning a larger dataset, training accuracy is often much higher than test accuracy. For example, ResNet18~\cite{ResNet} trained with CIRAR100~\cite{CIFAR100} achieves almost 100\% accuracy very quickly, but testing accuracy is between 72 to 75\%. This gap between training and test accuracy is viewed as (partial) memorization of (some) examples rather than learning underlying features and considered a generalization gap \cite{GeneralizationGap2022arXiv,FantasticGeneraizationMeasure2019arXiv}. A fundamental question is how to find what the network has learned for creating this gap. 
\par
Both theoretical and empirical approaches have attempted to understand the generalization phenomenon in DNNs (see \cite{FantasticGeneraizationMeasure2019arXiv} and the references therein).
A recent extensive study of thousands of different model trained with CIFAR10~\cite{CIFAR10} and SVHN~\cite{SVHN2011} datasets, evaluated with a wide range of carefully selected complexity measures (both empirical and theoretical), confirmed that {PAC}-Bayesian bounds provide good generalization estimates~\cite{FantasticGeneraizationMeasure2019arXiv}. Also, this study concluded that `several measures related to optimization are surprisingly predictive of generalization' but `several surprising failures about norm-based measures'. However, this study does not make efforts to estimate  how much accuracy a trained-DNN  would attain with `\emph{real}' test dataset, if no dataset is provided.

\par
 Assuming that a DNN is a composition of a feature extractor and a classifier, we show that the weight-vectors of a well-trained classifier are (almost) pairwise orthogonal 
(see Section\ref{sec:CharacteristicsOfTheFeatureExtracorHCdotThetaH}).
We propose two metrics for estimating quality of the feature extractor, which evaluates to `1'  with dataset of a network trained to neural collapse (see\cite{NeuralCollapse2020Journal} and Section~\ref{sec:trainingSetReconstraction2022a}); the metrics are defined and their computations methods are presented in Section~\ref{sec:TrainingQualityOfTheFeatureExtractor}.  We propose and evaluate a method for estimating quality of the feature extractor using these metrics. To be more precise, we  estimate test accuracy bounds of a trained DNN classifier using the proposed metrics.
\par
For computing values of the proposed metrics, an input dataset is necessary. If a dataset is provided, it would measure performance of the trained DNN, not training quality of the network. To overcome this, we propose a method for generating a necessary dataset from the trained DNN (without any data).
The first step of the proposed method is generation of one (input) prototype vector for each class (see Algorithm~\ref{algo:generateSeeds}). Then using these seed prototypes, $(k-1)$ \emph{core} prototypes are generated for each class (see Algorithm~\ref{algo:generateCorePrototypes}). These prototypes are our dataset for characterizing the DNN.

\par
After forward-passing each prototype through the feature extractor, a feature vector is obtained. Repeating this process with all the prototypes, a total of $k^2$ feature vectors, $k$ for each class, are generated and used to estimate mean values of two proposed metrics and their variances, which are used to produce an estimate of test accuracy of the given DNN. The metric for assessing the classifier, we use its weight vectors. For concise and precise descriptions we introduce the notions that we use in the paper.

\subsection{Notations} 
\label{Notations}   

\par
Let $[n]$ be the set $\{1,2,\cdots,n\}$.
Given,  $\mathcal{D} = \{(x_i,y_i) | x_i \in \R^p ,\, \, y_i \in [k] \,\,\mbox{and} \,\,i \in [n]\}$, a dataset of  $n$ labeled data coming from $k$ classes $C_1$ to $C_k$. Let $\mathcal{D}_l = \{(x_i, l)| (x_i,l) \in \mathcal{D} \,\, \mbox{and} \,\, l \in [k] \}$ be the set of data in classes $C_l$. Denoting $|\mathcal{D}_l| = n_l$, we have $n = \sum_{i=1}^k n_i$.
\par
 We consider multiclass classifier functions
$f: \R^p \rightarrow \R^k $ that maps an input $\vec{x} \in \R^p$  to a class $l \in [k]$.
For our purposes, we assume $f$ is a composition of a \emph{feature extraction} function $g:\R^p \rightarrow \R^q$ and a \emph{feature classification} function $h: \R^q \rightarrow \R^k$.  
Assume that $f$ is differentiable and after training with a dataset $\mathcal{D}$, a parametric function $f(\cdot\ ; \theta)$ is obtained.
Because $f$ is a composition of $g$ and $h$, training of  $f(\cdot\ ; \theta)$ produces $g(\cdot; \theta_g)$ and $h(\cdot; \theta_h)$, where $\theta = \theta_g \cup \theta_h$ and $ \theta_g \cap \theta_h = \oslash$.
\par
Let $\vec{v} = g(\vec{x};\theta_g)$ and $\vec{\hat{y}} = h(\vec{v};\theta_h)$,
where  $\vec{\hat{y}}_l = h_l(\vec{v};\theta_h)$ represents the probability 
$h_l(\vec{v};\theta_h)$ 
belongs to class $l \in [k]$ such that $\sum_{l=1}^{k} \hat{y}_l = 1$.  Correspondingly, for any given $\vec{x}$, which possess a true label $l$, the classifier could generate $k$ possible class assignments and is correct only when $l = \argmax_k(\vec{\hat{y}})$. 
\par
\emph{Unless stated otherwise}, we assume that activation functions at the output layer of the classifier $g(\cdot; \theta_g)$ are {ReLu}, and 
 $h(\cdot;\theta_h)$ is a fully connected one layer neural network, where $\vec{W}_{i,:}, i \in [k]$ be the $i$th row of $h(\cdot;\theta_h)$'s weight matrix $\vec{W}$.
It receives input from a feature extractor $g(\cdot; \theta_g)$, and $h(\cdot;\theta_h)$'s output is using a softmax function (see Eqn.\ref{SoftMax}) to generate one-hot coded outputs.
\begin{align}
	\sigma_l(\vec{W}\vec{v}) = \frac{e^{(\vec{W}_{l,:} \cdot \vec{v})}}{\sum_{i=1}^{k} e^{(\vec{W}_{i,:} \cdot \vec{v})}} \label{SoftMax}
\end{align}

\paragraph{Problem Statement}\label{Problem}

We are given a trained neural network $f(\cdot;\theta) = h(g(\cdot; \theta_{g});\theta_h)$ for the classification of input vectors $\vec{x} \in \R^p$ to one of the $k$ classes; that is, we are given the architecture of the network and values of the parameters $\theta$. No training, validation, or testing data is provided.The network uses a one-hot encoding using softmax function. The problem is to 
characterize training quality of the $f(\cdot;\theta) = h(g(\cdot; \theta_{g});\theta_h)$ without any test data.
 \emph{This is a harder problem, because no data is provided to us.}

\par
Our objective is to develop methods for assessing the network's training quality (accuracy). To be more specific, we want to develop metrics that can be used to estimate training quality of $g(\cdot;\theta_g)$ and $h(\cdot; \theta_h)$. We propose and empirically validate two metrics for $g$ and one for $h$. The metric for $h$ is based on the weight vector $\vec{W}$ of $h$, and metrics for $g$ are based on the features that
$g$ outputs when synthesized vectors are input to $g$. We propose a method for generating \emph{synthesized vectors}  and call them \emph{prototype vectors} (see Sec.~\ref{sec:MethodsForDataGeneration}) unless stated otherwise. 

\par
We propose a method for evaluating a trained network without knowledge of its training data and hyperparameters.
Our method involves \begin{inparaenum}[$(i)$] \item generation of \textbf{\emph{prototype data}} for each class, \item  observation of activations of neurons at the output of the feature extraction function $g(\cdot; \theta_g)$ (which is input to the classification function $h(\cdot; \theta_h)$), and \item asses quality of the given network using observations from previous step. \end{inparaenum} The last step, assessment of the quality of a trained network without testing data, requires one or more metrics, which are developed here. 

\par
The main contribution of our work are:
\begin{itemize}
  \item We have theoretically established that the weight-vectors of a well-trained classifier is (almost) orthogonal. Recall that we assume that a DNN is a composition of a feature extractor and a classifier.
	\item We have defined two metrics for estimating training quality of the feature extractor.  Using these two metrics we have developed a method for estimating lower and upper bounds for test accuracy of a given DNN without any knowledge of training, validation, and test data.
	\item We have proposed a method for generating a synthetic dataset from a trained DNN with one-hot coded output. Because each data in the dataset is correctly classified by the DNN with very high probability values, if the DNN is well trained the data should represent an input area from where the DNN must have received training examples.
	\item We have empirically demonstrated (almost) orthogonality of the weight vectors of the classifier section of
	ResNet18 trained with CIFAR-10, and CIFAR-100.
	\item We have empirically demonstrated that the proposed method can estimate lower and upper bounds of  actual test accuracy of ResNet18 trained with CIFAR-10, and CIFAR-100.
\end{itemize}

\par
The rest of the paper is organized as follows. Section~\ref{sec:CharacteristicsOfTheFeatureExtracorHCdotThetaH} studies characteristics of a well trained classier. It is proven that the weight vectors of the classifier are (almost) orthogonal. In Section~\ref{sec:TrainingQualityOfTheFeatureExtractor} introduces two metrics for measuring quality of training of a feature extractor. These two estimates upper and lower bounds of a feature extractor. Section~\ref{sec:MethodsForDataGeneration} proposes two algorithms for generating synthetic input images for characterizing a feature extractor. In Section~\ref{EmpiricalEvaluationOfDNNs}, we present some typical empirical results for validating theoretical result in the Section~\ref{sec:CharacteristicsOfTheFeatureExtracorHCdotThetaH} and for estimating training quality of feature extractors utilizing metrics in Section~\ref{sec:TrainingQualityOfTheFeatureExtractor} and synthetic data generation algorithms in Section~\ref{sec:MethodsForDataGeneration}.  Section~\ref{sec:trainingSetReconstraction2022a}  reviews work related to our work. In Section~\ref{Conclusion} we present concluding remarks and our efforts for improving the results.

\section{Characteristics of the Classifier  $h(\cdot;\theta_h)$ }
\label{sec:CharacteristicsOfTheFeatureExtracorHCdotThetaH}

An optimal representation of training data minimize variance of intra-class features and maximize inter-class variance \cite{RepresentationLearningAReview2013PAMI,LearningMoreUniversalRepresentationForTransferLearning2019PAMI}.  Inspired by these ideas in these hypotheses, we define training quality metrics  using cosine similarity of features of \emph{prototypes generated from trained DNNs} (see Sec~\ref{sec:MethodsForDataGeneration}). Let us deal with a much simpler task of defining a metric for a classifier $h(\cdot;\theta_h)$ defined as the last completely-connected layer.
\subsection{Training Quality of the Classifier}
\label{sec:TrainingQualityOfTheClassifier}

\par
Let  $\vec{W}_{i,:}$ be the $i$th row of $h(\cdot;\theta_h)$'s weight matrix $\vec{W}$. Recall that elements of the input vector $\vec{v}$ to $h(\cdot;\theta_h)$ are non-negative, because they are outputs of {ReLU} functions.

\begin{definition}[Well trained Classifier $h(\cdot;\theta_h)$]  \label{def:wellTainded} 
A one-hot output classifier $h(\cdot;\theta_h)$ with weight matrix $\vec{W}^{k\times q}$ is well trained, if for each class $l$ there exists a feature vector $\vec{v}^{(l)}$ such that
\begin{equation}
e^{(\vec{W}_{l,:}~\cdot~\vec{v}^{(l)})} >> e^{(\vec{W}_{i,:}~\cdot~\vec{v}^{(l)})}  \,\,\, \mbox{for all $i\neq l$, and } \label{WellTrainedCond1} 
\end{equation}
\begin{equation}
e^{(\vec{W}_{i,:}~\cdot~\vec{v}^{(l)})}  \leq 1 \,\,\,\, \mbox{for all $i\neq l$.} \label{WellTrainedCond2}
\end{equation}
\end{definition}

 For correct classification, the  condition~\ref{WellTrainedCond1} is sufficient. The condition~\ref{WellTrainedCond2} is a stronger requirement but enhances the quality of the classifier, and it leads us to the following lemma.


\begin{lemma}[Orthogonality of Weight Vectors of $h(\cdot;\theta_h)$]\label{ortho} Weight vectors $\vec{W}_{i,:}$ of a well trained $h(\cdot;\theta_h)$ are (almost) orthogonal.  \label{OrthoWeights}
\end{lemma}

\begin{proof}
From the condition~\ref{WellTrainedCond2} of the Def.~\ref{def:wellTainded} we assume, 
$e^{(\vec{W}_{i,:} \cdot \vec{v}^{(l)})} =1 \,\,\,\,\  \mbox{for all $i \neq l$}.$
This implies that $(\vec{W}_{i,:} \cdot \vec{v}^{(l)}) =0$ for all $i \neq l$, that is, $(k-1)$ weight vectors $\vec{W}_{i,:}$ and $i \neq l$ are orthogonal to $\vec{v}^{(l)}$. 
\par
Now the value of $(\vec{W}_{l,:} \cdot \vec{v}^{(l)})$ will determine the angle between $\vec{W}_{l,:}$ and  $\vec{v}^{(l)}$. Because of the condition~\ref{WellTrainedCond1} of the Def.~\ref{def:wellTainded}, it is much higher than zero and a maximum of
$(\vec{W}_{l,:} \cdot \vec{v}^{(l)}) = ||(\vec{W}_{l,:}||\,|| \vec{v}^{(l)})||$ is obtained, when they are parallel to each other. If maximum is reached then $\vec{W}_{l,:}$ is perpendicular to all other weight vectors. When this maximum is reached for all $l$, the weight vectors $\vec{W}$ are mutually orthogonal. Otherwise, they are almost orthogonal.
\end{proof}

Our empirical studies, reported in Sec.~\ref{EmpiricalEvaluationOfDNNs}, found that weight vectors of ResNet18 trained with CIFAR10 and CIFAR100  have average angles are 89.99 and 89.99 degrees, respectively. An obvious consequence of the Lemma~\ref{OrthoWeights} is that there exists a set of $k$  feature vectors  each of which are (almost) parallel to the corresponding weight vectors, which is stated as a corollary next.

\begin{corollary}[Parallelity of $\vec{W}_{l,:}$ and $\vec{v}^{(l)}$] 
If a classifier is well trained, for each weight vector $\vec{W}_{l,:}$ there is one or more feature vectors $\vec{v}^{(l)}$ that are (almost) parallel to it.
\end{corollary}

\paragraph{Classifier's Quality Metric, $\mathcal{H}_w$,}
\label{sec:ClassifierSQualityMetric}
 postulates that weight vectors of a well trained classifier are (almost) orthogonal to each other. A method for its estimation is defined next.

\begin{definition}[Estimation of metric $\mathcal{H}_w$] An estimation of  $\mathcal{H}_w$ is calculated by consider a pair-wise combinations of the weight matrix of $h$, and given by
\begin{equation} \label{eqn:WtMetric} 
	\hat{\mathcal{H}}_{w} = 
	1-\frac{2}{k(k-1)}\sum_{j<i\leq l} \left( \frac{\vec{W}_{i,:} \cdot \vec{W}_{j,:}}{||\vec{W}_{j,:}||\,||\vec{W}_{j,:}||}.
	\right)
\end{equation}
\end{definition}

A well trained classifier will have a value of $\hat{\mathcal{H}}_{w}$ very close to 1.
\par
\textbf{Note} that $k$ weight vectors are projecting a $q$-dimension feature vector to a $k$-dimension vector space, for $q > k$. If the weight vectors are orthogonal, their directions define an orthogonal basis in the $k$-dimension space. Because of the isomorphism of the Euclidean space, once the weight vectors of classifier $h(\cdot;\theta_h)$ are (almost) orthogonal their updating is unlikely to improve overall performance of the DNN and at that point, the focus should be to improve the feature extractor's parameters.
\par
In the next section, we develop a method for evaluation of training quality of the feature extractor $g(\cdot;\theta_g)$.
It is a much harder task because feature extractors are, in general, much more complex and \emph{directly quantifying} its parameters with a few metrics is a near impossible task; we  develop a method for \emph{indirectly evaluating} the feature extractor without training or testing data.
\section{Evaluataion of Feature Extractor $g(\cdot;\theta_g)$}
\label{sec:EvaluataionOfFeatureExtractorGCdotThetaG}

\subsection{Evaluation Metrics for the Feature Extractor $g(\cdot;\theta_g)$}
\label{sec:TrainingQualityOfTheFeatureExtractor}  


For evaluation of a feature extractor $g(\cdot; \theta_g)$ we use a labeled data set $\mathcal{D}^{E}$ with $|\mathcal{D}^E| = n^E$.  
We assume, for simplicity of exposition, each evaluation class has the same number of examples making $n_l^E = n^E/k$.
Let  $\mathcal{V}^E_l$ be the set of feature vectors generated by
$g(\cdot;g_{\theta })$ from all evaluation data in $\mathcal{D}^E_l$. 
We  define two metrics for measuring performance of the feature extractor $g(\cdot;g_{\theta })$ with the evaluation data $\mathcal{D}^L$: one for within class training quality and the other for between class training quality. Within class features of a class is considered perfect, when all feature vectors for a given class are pointing in the same direction, which is determined by computing pair-wise cosine of the angle between them. We discuss between class training quality later. Let us formally define cosine similarity.

\begin{definition}[Cosine Similarity] Given two vectors $\vec{v}_1$ and $\vec{v}_2$ of same dimension, cosine similarity 
\begin{equation} \label{def:CosSim} 
CosSim(\vec{v}_1, \vec{v}_2)) = \cos(\theta) \equiv (\vec{v}_1 \cdot \vec{v}_2)/(||\vec{v}_1||\,||\vec{v}_2||)
\end{equation}
\end{definition}

Computationally, $CosSim(\vec{v}_1, \vec{v}_2)$ of $\vec{v}_1$ and $\vec{v}_2$ is the  \emph{dot} product of their unit vectors. 
We normalize all vectors in $\mathcal{V}^E_{l}$  
for ease of computing all-pair $CosSim$ of the vectors in it. For a set of $n^E_l$ vectors in each class, we need to compute $n^E_l(n^E_l-1)/2$ $CoSim$ since that is the number of possible pairwise combinations. Matrix multiplication  provides a convenient method for computing these values.
Let $\vec{v}^E_{l,i}$ be the $i$th row of the matrix $\vec{G}^{(l)}$ and let $(\vec{G}^{(l)})^T$ be the transpose of $\vec{G}^{(l)}$. 
The elements of the $k\times k$ matrix $(\vec{G}^{(l)} (\vec{G}^{(l)})^T)$ are 
the 
$CosSim(\vec{v}_{c,i}^{(l)}, \vec{v}_{c,j}{(l)})$. 
 Each diagonal element, being \emph{dot} products of a unit vector with itself, is `one' and because
$(\vec{G}^{(l)} (\vec{G}^{(l)})^T)_{i,j}$ =$(\vec{G}^{(l)} (\vec{G}^{(l)})^T)_{j,i}$,  we need to consider only off-diagonal the upper or lower triangular part of the matrix.

\subsubsection{Within-Class Similarity Metric}
\label{sec:WithinClassSimilarityMetric}

\paragraph{Intra-class or within-class Similarity Metric,}$\mathcal{M}_{in}$, postulates \emph{intra-} or \emph{within-}class 
features
of a well trained DNN are very similar  for a given class and their pairwise $CosSim$ values are close to \emph{one}. 
To get an estimate within-class similarity of a DNN, estimates of all classes are combined together.


\begin{definition} [Estimate of intra- or within-class similarity metric]
An estimate of intra- or within-class similarity metric $\mathcal{M}_{in}$ is given by
\begin{equation} \label{eqn:MsMetric} 
	\hat{\mathcal{M}}_{in} = 
	\frac{1}{k}\sum_{l\in {[k]}} \left(\frac{2}{n_l^E(n_l^E-1)}\sum_{{j < i \leq n_l^E}}((\vec{G}^{(l)} (\vec{G}^{(l)})^T)_{i,j})\right)
\end{equation}
\end{definition}

 The variance of $\mathcal{M}_{in}$ of  a well trained DNN should be very low as well. Our evaluations reported in Sec.~\ref{EmpiricalEvaluationOfDNNs} show that $\hat{\mathcal{M}}_{in}$ for CIFAR10 is above 0.972 even when it is  trained with 25\% of the  trained data and variance is also small, an indication that the DNN was probably trained well. But for CIFAR100 $\hat{\mathcal{M}}_{in}$ and variance are relatively higher, indicating that it was not trained well.
\emph{We propose to use within-class similarity metric to define upper bound for test accuracy} (see Sec.~\ref{EmpiricalEvaluationOfDNNs}).

\subsubsection{Between-Class Separation Metric}
\label{sec:BetweenClassSeparationMetric}

\paragraph{Inter-class or between-class Separation Metric,}$\mathcal{M}_{bt}$, postulates that two prototypes of different classes should be less similar and their pair-wise $CosSim$ value should be close to \emph{zero}. For making $\mathcal{M}_{bt}$ increase towards one as the quality of a trained DNN increases, we define $\mathcal{M}_{bt} = 1- CosSim(\vec{v}_{c,i}^{(l_1)}, \vec{v}_{c,j}^{(l_2)})$, where $\vec{v}_{c,j}^{(l_1)}$ and $\vec{v}_{c,j}^{(l_2)}$ are feature vectors from two different classes $l_1$ and $l_2$,  respectively.  
\par
Because every class $l\in {[k]}$ has $n_l^E$ evaluation examples, estimation of $\mathcal{M}_{bt}$ requires further considerations. Consider two classes $l_1$ and $l_2$, we can pick one evaluation example from class $l_1$ and compare it with $n_{l}^E$ examples in $l_2$ (recall that we assumed that all classes have same number evaluation examples); repeating this process for all examples in the class $l_1$, require $O((n_{l}^E)^2)$ $CosSim$ computations. Also, because each example in class $l_1$ must be compared with all examples in other $(k-1)$ classes, total $O((k-1)(n_l^E)^2)$ $CosSim$ computations are necessary. 
Overall total computational complexity for all $k$ classes is $O(n^E)^2$. If computation time is an issue, one can  get a reasonable estimate with $O((n^E)^2/k)$ computations by comparing a mean feature vector of a class $l$ with all vectors of remaining of $(k-1)$ classes. To reduce computations further, one can compute $k$ mean vectors, one for each class and them compute lower (or upper) triangular part of $(k\times k)$ $CosSim$ matrix requiring a total of $O(k(n_l^E + k-1)/2))$ computations. We will define $\hat{\mathcal{M}}_{bt}$ using the 2nd case, where for a class $l \in {[k]}$ we compute average feature vector $\hat{v}_l^E$ from feature vectors 
 $g(\cdot;g_{\theta })$ generates from all examples in $\mathcal{D}^E_{l}$.

Assume that $g(\cdot;g_{\theta })$ generates $n_{l_i}^E$ feature vectors $\mathcal{V}^E_{l_i}$  from all evaluation data in $\mathcal{D}^E_{l_i}$, which are
normalized to $v_{l_i,1}^{E}, v_{l_i,2}^{E},\cdots, v_{l_i,n_{l_i}}^{E}$ and a matrix $\vec{H}_{l_i}$ is created, where $v_{l_i,j}^{E}$ is the $j$th row of it.



\begin{definition} [Estimate of inter- or between-class separation metric]
An estimate of inter- or between-class separation metric $\mathcal{M}_{bt}$ is given by
\begin{equation} \label{eqn:BtMetric} 
	\hat{\mathcal{M}}_{bt} = 
	1-\frac{1}{k(k-1)}\sum_{l \in {[k]}}\sum_{\substack{{l_i\in {[k]}}\\{l \neq l_i}}} \left(\frac{1}{n_{l}^E }\sum_{j =1}^{n_{l}^E}(\vec{H}_{l_i,j}[\hat{v}_l^E]^T)\right)
\end{equation}
\end{definition}

\subsection{ Training Quality of Feature Extractor }
\label{sec:EvaluationOfTrainingQualityOfDNNs}

When a feature classifier's weights are pairwise orthogonal, 
overall classification accuracy will depend on the features that the feature extractor generates. If feature vectors of all testing examples of a class are similar and their standard deviation is very low, the value of $\hat{\mathcal{M}}_{in}$ will be high; but features of the between-class  are very dissimilar  and their standard deviation is high, the value of $\hat{\mathcal{M}}_{bt}$ will be low. This situation may cause classification  errors, when feature vector of an example is too far from $\hat{\mathcal{M}}_{in}$. Our empirical observations is that  during training time within-class feature vectors' become very close to each other faster and this closeness represents an upper bound for testing accuracy.    
\par
Also, we observed that it is much harder to impart training so that  feature vectors of one class move far away from that of other classes, which keeps the value of the metric $\hat{\mathcal{M}}_{bt}$ lower, and hence, it will act as a lower bound for the testing accuracy. This hypotheis is supported by our empirical observations support.
\par
 In the next section, we empirically demonstrate that estimates of $\mathcal{M}_{in}$  and $\mathcal{M}_{bt}$ are proportional to test accuracy. Moreover,  estimated values of $\mathcal{M}_{in}$ and $\mathcal{M}_{bt}$  serve as predictions of the upper- and lower-bounds of the test accuracy of the network. 

\subsubsection{Evaluation of Training quality of DNNs without Testing Data}
\label{sec:EvaluationOfTrainingQualityOfDNNsWithoutTeastingData}
The metrics defined above and methods for estimation of their values for a network requires data, which  are used for generating feature vectors. Thus, 
estimation of  within-class similarity and between-class separation metrics of a trained DNNs  can be a routine task when testing data is available. But can we evaluate a DNN when no data is available? We answer this affirmatively.

In the next section, we propose a method for generation of evaluation examples from a DNN itself. 

\section{Data Generation Methods for Feature Extractor Evaluation}
\label{sec:MethodsForDataGeneration}

We employ the provided DNN to \emph{synthesize} or \emph{generate} input data vectors, which are then utilized to evaluate the feature extractor. We use the terms `synthetic' or `generated' interchangeably. These generated data are referred to as \emph{prototypes} to differentiate them from the \emph{original} data used in developing the DNN.
The generation methods described next ensure that the synthesized data is correctly classified.
\par
 A prototype generation starts with an random input image/vector and a target output class vector. 
he input (image) undergoes a forward pass to produce an output. The output of the DNN and the target class are then used to compute loss, which is subsequently backpropagated to the input. The input image is updated iteratively until the loss falls below a predetermined threshold. Since the synthesized data is generated using the DNN's parameters, it reflects the inherent characteristics of the trained DNN. Below, we outline our proposed data generation methods.

\subsection{Generation of Prototype Datasets}
\label{sec:SeedDataset}

To iteratively generate a prototype, we employ the loss function defined by Eqn. \ref{lossFunction1} (although this is not a limitation, as the method can be used with other loss functions as well) and the update rule described in Eqn. \ref{updateRule1}, where $t$ represents the iteration number and $\vec{m}_t$ denotes the current input to the network.
\begin{align}
		\textrm{\textbf{Loss Function: }}&\mathcal{L}  = - \sum_{j=1}^{k} y_{\vec{m}_t,j}\ log(f_j(\vec{m}_t)) 
		\label{lossFunction1}
		\end{align}
		\begin{align}
	\textrm{\textbf{Update Rule: }} &\vec{m}_{t+1} \leftarrow \vec{m}_{t} - \eta\nabla_{\vec{m}_t}\mathcal{L}\ / ||\ \nabla_{\vec{m}_t}\mathcal{L}\ || \label{updateRule1} 
\end{align}

Let us consider the one-hot encoded output vector $\vec{y}_l$, where for a particular class $l$ belonging to the set $[k]$, the element $y_l$ is set to 1 to indicate membership in that class, while all other elements $y_j$ for $j$ not equal to $l$ are set to 0 to signify non-membership in those classes.  

To generate a prototype example $\vec{m}^{(l)}$ for class $l$, we begin by initializing $\vec{m}^{(l)}_0$ with a random vector $\vec{r} \in \mathbb{R}^p$. We then perform a forward pass with $\vec{m}^{(l)}_0$ to compute the cross-entropy loss between the output $f(\vec{m}^{(l)}_0;\theta)$ and the one-hot encoded target vector $\vec{y}_l$. Subsequently, we backpropagate the loss through the network and update the input vector using the rule described in Eqn. \ref{updateRule1} to obtain $\vec{m}^{(l)}_1$. This iterative process continues until the observed loss falls below a desired (small) threshold.

\textbf{Probabilities for Core Prototypes:}

A defining feature of a core prototype for a class $l \in [k]$ lies in its capacity to produce an optimal output. In essence, the output probability $p_l$ corresponding to the intended class ideally tends towards `1', reflecting a high level of confidence in the classification. Conversely, the probabilities associated with all other classes ideally tend towards `0', denoting minimal likelihood of membership in those classes. This characteristic ensures that the prototype adeptly encapsulates the unique attributes of its assigned class, thereby facilitating precise and reliable classification.

\begin{align}
p_l = 1\,\,\,  \mbox{and}\,\,\, p_j = 0 \,\,\,\mbox{for}\,\,\, j\neq l \label{saturatingProb}
\end{align}

 First, we use the above probability distribution to generate a set of \emph{seed} prototypes, which are then used to generate core prototypes (see description below). 
\subsection{Generation of Seed Prototypes}
\label{sec:SeedDataseta} 

First we generate a set of $k$ \emph{seed} prototypes, exactly one seed for each category. The starting input vector for a seed prototype is a random vector $m_0 \in \R^{p}$, a target class label $l \in [k]$ and a small loss-value $\delta_{loss}$ for termination of  iteration (see Sec.~\ref{sec:SeedDataset}). We use these  seed prototypes  to generate $(k-1)$ \emph{core prototypes}  for each category.

\begin{algorithm}[htb]
	\caption{Generate $k$ Seed Prototypes $\vec{S}$}
	\label{algo:generateSeeds}
	\begin{algorithmic}[1]
		\Procedure {GenerateSeedsPrototypes} {} 
		\State Input: $\delta_{loss}$ \Comment Loss for iteration termination 
		\State Output: $\vec{S}$
		\State $\vec{S} = \oslash$
		\State for each $ l \in k$ 
		\State \hsb $\vec{S}_0^{(l)} = \vec{r} \in \R^p$, $\mathcal{L} =\infty$, and $t = 0$
		\State \hsb while $\mathcal{L} > \delta_{loss}$ 
		\State \hsd $\mathcal{L}  = - \sum_{j=1}^{k} y_{\vec{S}_t^{(l)},j}\ log(f_j(\vec{S}_t^{(l)}))$ \Comment Eqn. \eqref{lossFunction1}
		\State \hsd $\vec{S}_{t+1}^{(l)} = \vec{S}_{t}^{(l)} - \eta\nabla_{\vec{S}_{t}^{(l)}}\mathcal{L}\ / ||\ \nabla_{\vec{S}_{t}^{(l)}}\mathcal{L}\ ||$  \Comment Eqn.~\eqref{updateRule1}
		\State \hsd $t = t +1$
		\State \hsb $\vec{S} = \vec{S} \bigcup\vec{S}_{t}^{(l)}$
		\EndProcedure
	\end{algorithmic}
\end{algorithm}

Let $\vec{S}_0 = \{ \vec{S}_{0}^{(l)}| l \in [k] \}$ be the set of $k$ random vectors drawn from $R^p$. Starting with each $\vec{S}_{0}^{(l)}$ generate one seed prototype for class~$l$ by iteratively applying update rule given by  Eqn.~\ref{updateRule1} and output probability distribution defined by Eqn.~\ref{saturatingProb}. Let the seed vector obtained after termination of iterations that started with initial vector $\vec{S}_{0}^{(l)}$  be denoted by $\vec{S}^{(l)}$ and let $\vec{S}$ be the collection of all such $k$ vectors. A procedure for generating $n$ seed-prototypes is outlined in the Algorithm \ref{algo:generateSeeds}.
\par
These seed vectors  serve two purposes: a \emph{preliminary} evaluation of inter-class separation and \emph{initial} or \emph{starting input} for generating prototypes to characterize intra-class compactness. We generate $(k-1)$ prototypes for each class.

\begin{algorithm}[htb]
	\caption{Generate $k(k-1)$ Core Prototypes $\vec{S}_C$}
	\label{algo:generateCorePrototypes}
	\begin{algorithmic}[1]
		\Procedure {GenerateCorePrototypes} {} 
		\State Input: $\delta_{loss}$ and $S$ \Comment Loss for iteration termination and $k$ seed prototypes
		\State Output: $\vec{S}_c$ \Comment $S_c$ will have $k(k-1)$ core prototypes, $(k-1)$ for each class
		\State $\vec{S}_c = \oslash$
		\State for each $ l \in k$ 
		\State \hsb for each $ (j \in k)$ and $j \neq l$ \Comment Find $(k-1)$ prototypes for class $l$
		\State \hsc $\vec{S}_{c,0}^{(j)} = \vec{S}^{(j)}$, $\mathcal{L} =\infty$, and $t = 0$ \Comment Start from the seed prototype in the class $j$
		\State \hsc while $\mathcal{L} > \delta_{loss}$ 
		\State \hse Compute $\mathcal{L}$ using probability distribution given by Eqn. \eqref{saturatingProb} for class $l$
		\State \hse $\vec{S}_{c,t+1}^{(j)} = \vec{S}_{c,t}^{(j)} - \eta\nabla_{\vec{S}_{c,t}^{(j)}}\mathcal{L}\ / ||\ \nabla_{\vec{S}_{c, t}^{(j)}}\mathcal{L}\ ||$  \Comment Eqn.~\eqref{updateRule1}
		\State \hse $t = t +1$
		\State \hsb $\vec{S}_c = \vec{S}_c \bigcup \vec{S}_{c,l}^{(j)}$
		\EndProcedure
	\end{algorithmic}
\end{algorithm}

\subsection{Generation of Core or Saturating Prototypes}
\label{sec:CoreOrSaturatingPrototypes}

 Let $S^{(l)}_{c,i}$ be the prototype for class $l$ generated starting from the seed  $S^{(i)}$ of the class $i$. We expect that $S^{(l)}_{c,i}$ to be the closest input data that is within the boundary of the class $l$ and closest to the boundary of the class $i$.
It is important to note that output produced by prototype $\vec{S}^{(j)}$ satisfies the Eqn.~\ref{saturatingProb}. As shown in the Algorithm~\ref{algo:generateCorePrototypes}, the generation process initiates a prototype with $\vec{S}^{(j)}, j \neq l$ and iterates using update rule (Eqn.~\ref{updateRule1}) until it produces outputs that satisfy Eqn.~\ref{saturatingProb} for the class $l$. Basically, we are starting with a prototype within the polytope of the class $j$ and iteratively updating it until the updated prototype has moved (deep) inside the class $l$. 
We repeat the process for all $l \in [k]$ to generate $\vec{S}_c = \{\vec{S}_c^{(1)}, \vec{S}_c^{(2)}, \cdots, 
\vec{S}_c^{(k)}\}$.
Since the output produced by these prototypes create saturated outputs, we call the prototypes in $\vec{S}_c$ \emph{core} or \emph{saturating} prototypes.


The prototypes generated using two algorithms described in this section are used for evaluation of the quality of the feature extractor.

\section{Empirical Evaluation of DNN Classifiers}
\label{EmpiricalEvaluationOfDNNs}

\paragraph{Datasets}
We have used the image classification datasets CIFAR10 \cite{CIFAR10}, and CIFAR100 \cite{CIFAR10}.
All of these datasets contain labeled 3-channel 32x32 pixel color images of natural objects such as animals and transportation vehicles. CIFAR10 and CIFAR100 have  10 and 100   categories/classes, respectively. The number of training examples per category  are 5000 and 500 for CIFAR10 and 100, respectively. 
\emph{We created 7 datasets from each original dataset by randomly selecting 25\%, 40\%, 60\%, 70\%, 80\%, 90\%, and 100\% of the data}. 

\paragraph{Training}
For results reported we have used ResNet18 \cite{ResNet}, defining $g(\ \cdot\ ;\theta_{g})$ as the input to the flattened layer (having 256 neurons) after global average pooling and $h(\ \cdot\ ;\theta_{h})$ as the fully connected and softmax layers.  All evaluations were completed on a single GPU.
\par
For each dataset, we randomly initialize a ResNet18 \cite{ResNet} network and train it for a total of 200 epochs. The learning rate starts at 0.1 in the first 100 epochs, and then it is reduced to 0.05 for the last 100 epochs. 
For each data set we trained 35 networks, 7 partitioned datasets and  5  randomly initialized weights to reduce biases.

\subsection{Evaluation of the Classifier}
\label{sec:EvaluationTrainingQualityOfTheClassifier}
In Section~\ref{sec:TrainingQualityOfTheClassifier} we have proved that a well trained classifier's weight vectors are  (almost) orthogonal. Here we empirically validate the Lemma~1.
With the DNN classifiers trained with CIFAR 10 and CIFAR 100 and then their weights were frozen.  Values of the  metric ${\hat{\mathcal{\vec{H}}}_w}$ were calculated using the weights of the last layer of the trained networks. Because the values were so close to one we converted them into angles and they are summarized in Table~\ref{Tab:OrthogonalWeightVectors}. It shows that the weight vectors are almost orthogonal even when only 25\% of the data is used.
\begin{table}[htb]
\small
\centering
\begin{tabular}{|l|l|l|l|l|l|l|l|}\hline
Taring Data \% & 25    & 40    & 60    & 70    & 80    & 90    & 100    \\\hline
CIFAR10        & 89.98 & 89.99 & 89.99 & 89.99 & 89.99 & 89.99 & 89.99 \\\hline
CIFAR100       & 89.99 & 89.99 & 89.99 & 89.99 & 89.99 & 89.99 & 89.99  \\\hline
\end{tabular}
\vspace{.1in}
\caption{Empirical validation of Lemma~1 for ResNet18; mean angles in degrees between weight vectors $\vec{W}$ of the Classifier $h(\cdot;\theta_h)$ are almost 90 degrees.}
\label{Tab:OrthogonalWeightVectors}
\end{table}
\normalsize

Next  we present evaluation results for the feature extractor.

\begin{table}[htb]
\small
\centering
\begin{tabular}{|l|ll|l||ll|l|l||l|} \hline
Taring  & Mean  &Within& $(\hat{\mathcal{M}}_{in}$ -  & Mean   &Between & CSbt &   & Accu- \\
Data \% &within &CosSIn &2*Std)& between &CosSim  & +2btStd &  $\hat{\mathcal{M}}_{bt}$   & racy \\
        &CosSim & Std   & $\hat{\mathcal{M}}_{in}$      & CosSim&Std     &  &   &  \\ 
				&$\hat{\mathcal{M}}_{in}$ &  (inStd) &   &(CSbt) &(CSbtStd)&&  &  \\ \hline 
25             & 0.9727                   & 0.0129                    & 0.9470     & 0.2354                    & 0.3922                      & 0.6002      & 0.6078 & 0.7099   \\ \hline
40             & 0.9793                   & 0.0093                    & 0.9607     & 0.2511                    & 0.3975                      & 0.5807      & 0.6025 & 0.7675   \\ \hline
60             & 0.9825                   & 0.0087                    & 0.9652     & 0.2277                    & 0.3585                      & 0.5292      & 0.6415 & 0.8149   \\ \hline
70             & 0.9837                   & 0.0082                    & 0.9673     & 0.2218                    & 0.3487                      & 0.5168      & 0.6513 & 0.8249   \\ \hline
80             & 0.9824                   & 0.0088                    & 0.9647     & 0.2339                    & 0.3697                      & 0.5113      & 0.6303 & 0.8362   \\ \hline
90             & 0.9849                   & 0.0072                    & 0.9704     & 0.2269                    & 0.3631                      & 0.4900      & 0.6369 & 0.8449   \\ \hline
100            & 0.9844                   & 0.0077                    & 0.9690     & 0.2292                    & 0.3690                      & 0.4614      & 0.6310 & 0.8548 \\ \hline 
\end{tabular}
\vspace{.1in}
\caption{ Results for ResNet18 with 256 neurons at the feature layer and trained with CIFAR10.}
\label{tab:CIFAR10}
\end{table}
\normalsize
\subsection{Evaluation of the Feature Extractor}
\label{sec:EvaluationOfTheFeatureExtractor}
\par
For evaluating the feature extractor, we used the frozen networks to generate seed prototype images using Algorithm~\ref{algo:generateSeeds} with a learning rate of 0.01 for CIFAR10 and 0.1 for CIFAR100, respectively. Then Algorithm~\ref{algo:generateCorePrototypes} was used to generate $(k-1)$ \emph{core} prototypes (see Sec.\ref{sec:CoreOrSaturatingPrototypes}) for each category. Thus, for CIFAR10 we generate a total of 100 prototypes and for CIFAR100 we generate a total of 10,000 prototypes. The process was repeated five times to eliminate random selection bias and reported results are averages of these five runs. It is worth mentioning that we examined the data from each run and found no glaring differences.

%

\par
Our observations are summarized in Tables~\ref{tab:CIFAR10}~and~\ref{tab:CIFAR100}. Table~\ref{tab:CIFAR10} shows results for the CIFAR10 dataset and its fractions. The 7 rows are for the 7 datasets we created from the original dataset by partitioning it. The 1st column shows percent of data used. The second and third columns show values of the mean and standard deviation of the within-class similarity 
measure $\hat{\mathcal{M}}_{in}$. Column 8 shows that as the training dataset size increases, so does the testing accuracy.

\paragraph{An Upper Bound for Testing Dataset Accuracy:}
\label{UpperBoundMin}

The column~4 shows adjusted values of $\hat{\mathcal{M}}_{in}$, which is obtained by subtracting two standard deviations for increasing confidence of the observations. \emph{These values give us an upper bound for accuracy that one would have observed from the testing dataset.} As discussed before, for a given class as the value of $\hat{\mathcal{M}}_{in}$ increases and its standard deviation decreases, the features of the class are close to each other and test data performance should be high.

          \begin{figure}[htb]
			\includegraphics[width=.8\textwidth]{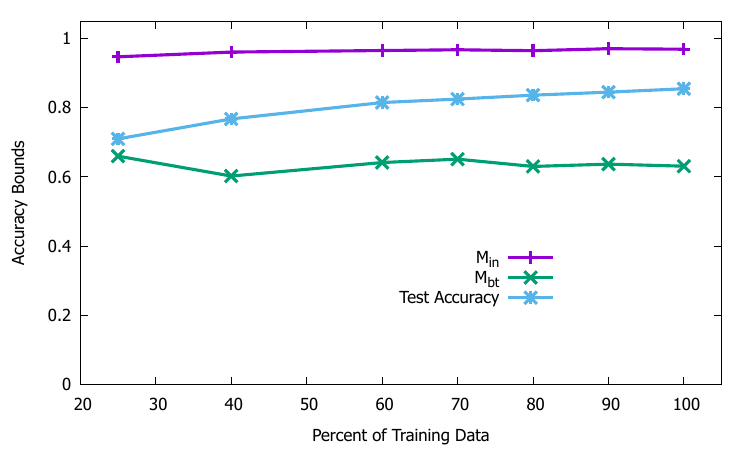}
          \caption{Accuracy bounds for ResNet18 with 256 neurons at the feature layer and trained with CIFAR10.}
          \label{fig:CIFAR10Accuracy}
          \end{figure}

\par
Let us explore the columns 5 to 8 of the Table~\ref{tab:CIFAR10}. Values in the columns 5 to 7 are analogous to the values in the values in the columns 2 to 4, but for between-class similarity measures. For a well trained network these values should be small. Calculation of values in column 4 differs from that in column 7 because it is obtained by adding (instead of subtracting) two standard deviations for increasing confidence of the observations.

\paragraph{A Lower Bound for Test Dataset Accuracy: }
\label{sec:ALowerBoundForTestDatasetAcuuracy}
The column 8 is obtained after subtracting the values in column 7 (see Eqn.~\ref{eqn:BtMetric}). \emph{These values give us a lower bound for accuracy one would get from the testing dataset.} As discussed earlier, for a given class as the value of $\hat{\mathcal{M}}_{bt}$ increases and its standard deviation decreases, the features of the class are much different from those of other classes and accuracy observed from testing dataset should increase.

\par
The data in the \emph{column 9 shows accuracy} of  the trained networks obtained \emph{when tested with test a dataset}. Note that we used all data in the test dataset (not a fraction as used for the training of the network). For ease of comparing and contrasting, the values of the Upper and Lower Bounds as well as test Accuracy for CIFAR10 is shown in Fig.~\ref{fig:CIFAR10Accuracy}. We can see that observed accuracy is correctly bounded by calculated values of the metrics.

\begin{table}[htb]
\small
\centering
\begin{tabular}{|l|ll|l||lll|l||l|} \hline
Taring  & Mean  &Within& $(\hat{\mathcal{M}}_{in}$ -  & Mean   &Between & CSbt &   & Accu- \\
Data \% &within &CosSIn &2*Std)& between &CosSim  & +2btStd &  $\hat{\mathcal{M}}_{bt}$   & racy \\
        &CosSim & Std   & $\hat{\mathcal{M}}_{in}$      & CosSim&Std     &  &   &  \\ 
				&$\hat{\mathcal{M}}_{in}$ &  (inStd) &   &(CSbt) &(CSbtStd)&&  &  \\ \hline 
25             & 0.9134 & 0.0285  & 0.8565     & 0.3721  & 0.1140    & 0.6002      & 0.3998 & 0.3256   \\\hline
40             & 0.9168 & 0.02523 & 0.8664    & 0.3656   & 0.1075      & 0.5807      & 0.4193 & 0.414    \\\hline
60             & 0.9233   & 0.0229   & 0.8776 & 0.3337    & 0.0977      & 0.5292      & 0.4708 & 0.4982   \\\hline
70             & 0.9248  & 0.0221   & 0.8807    & 0.3269   & 0.0949   & 0.5168      & 0.4832 & 0.5282   \\\hline
80             & 0.9252   & 0.0218  & 0.8815    & 0.3267   & 0.0923   & 0.5113      & 0.4887 & 0.5526   \\\hline
90             & 0.9273  & 0.0216   & 0.8842    & 0.3112     & 0.0894    & 0.4900      & 0.5100 & 0.5685   \\ \hline
100            & 0.9331   & 0.0204   & 0.8923 & 0.2897    & 0.0858                    & 0.4614      & 0.5386 & 0.5945  \\\hline
\end{tabular}
\vspace{.1in}
\caption{ Results for ResNet18 with 256 neurons at the feature layer and trained with CIFAR100.}
\label{tab:CIFAR100}
\end{table}

\normalsize

          \begin{figure}[h]
			\includegraphics[width=.85\textwidth]{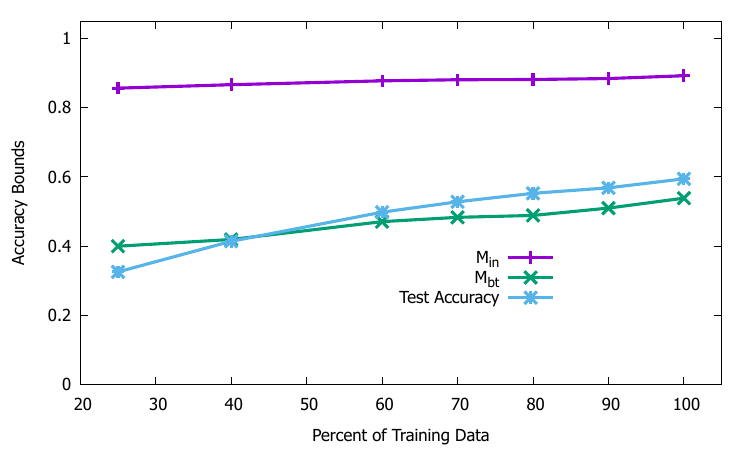}
          \caption{Accuracy bounds for ResNet18 with 256 neurons at the feature layer and trained with CIFAR100.}
          \label{fig:CIFAR100Accuracy}
          \end{figure}



\par

The Table~\ref{tab:CIFAR100}  and the Fig.~\ref{fig:CIFAR100Accuracy} are for CIFAR100 dataset. Except for 25\% and 40\% of the training data the bounds obtained from the network have enclosed the accuracy correctly. But a closer examination of the data in the table, tells us that 25\% and 40\% data created a very poorly trained network. Thus, it may not be worth using them for any serious classification applications.
\section{Related Work}
\label{sec:RelatedWork}
To the best of our knowledge, no work for estimation of testing accuracy of a trained DNN classifier without dataset exists, except in~\cite{fantasticClassifiers2023arXiv} where it was shown that test accuracy is correlated to cosine similarity of between class prototypes, but their prediction was inaccurate. Learning from their limitations, we propose a method for providing lower and upper bounds of test accuracy.
\par The phenomenon of neural collapse occurs for $L_2$ regularized neural networks that are trained for a sufficient epochs beyond zero training error. As the neural network trains towards zero classification loss the feature vectors at the penultimate layer of a class become parallel to each other and those between classes become perpendicular to each other.

\par 
A related and very important problem, generalization, has been studied theoretical and empirical by many. A good summary of these efforts can be found in \cite{FantasticGeneraizationMeasure2019arXiv}. After extensive evaluation of thousands of trained networks, \cite{FantasticGeneraizationMeasure2019arXiv}
reported that {PAC}-Bayesian bounds are good predictive measures of generalization and several optimizations are predictive of generalization, but regularization methods are not. Their evaluation method, like ours, doesn’t require a dataset, but their objective is \emph{generalization ability} prediction and our objective is \emph{test accuracy} estimation.
\par 
Since acquisition of labeled examples is expensive, active testing~\cite{ActiveTesting2021Kossen} and active surrogate estimator~\cite{ActiveSurrogateEstimator2022Kossem} selects samples to reduce the number of examples to reduce testing cost. However, it require testing samples.
\paragraph{Training Set Reconstruction:}
\label{sec:trainingSetReconstraction2022}
Recent work \cite{buzaglo2023,haim2022} has successfully reconstructed training examples by optimizing a reconstruction loss function that requires only the parameters of the model and no data.  An underlying hypothesis of this work is that neural networks encode specific examples from the training set in their model parameters. We utilize a similar concept to construct `\emph{meta}' training examples by optimizing a cross-entropy loss at the model's output and requiring significantly lower computation time.
\par
\paragraph{Term Prototype in Other Contexts:}
\label{sec:trainingSetReconstraction2022a}
We want the reader to be aware that term prototype we use have no \textbf{direct} relation to the training data and \textbf{it should not} be confused with or associated with same term has been used in many other contexts \cite{mensinkDBIC,rippelMagnet,snellPrototypeNetworksFewShot,liPrototype,karlinskyDML,chenProtoPNet,guerrieroDNCMC,mustafaRestrictedHiddenSpace}.

\section{Conclusion}
\label{Conclusion}
In this paper we have proposed and evaluated a method for dataless evaluation of trained DNN  that uses one-hot coding at the output layer, and usually implemented using softmax function. We assumed that the network is a composition of a feature extractor and a classifier. The classifier is a one-layer fully-connected neural network that classifies features produced by the feature extractor. We have shown that the weight vectors of a well trained classifier-layer are (almost) orthogonal. Our empirical evaluations have shown that the orthogonality is achieved very quickly even with a small amount of data and test accuracy of the overall DNN is quite poor. 
\par
The feature extractor part of a DNN is very complex and has a very large number (usually many millions) of parameters and their direct evaluation taking into consideration all parameters is an extremely difficult, if not impossible, task. We have proposed two metrics for indirect estimation of feature extractor's quality.
Values of these metrics are computed using activation of neurons at the output layer of the feature extractor. 
For observing neuron activations we generate synthetic input data using the DNN classifier to be evaluated. 
\par 
After feeding synthetic input data we record feature extractor's responses and compute values of the two metrics, which are  estimates for an upper and lower bound for classification accuracy. Thus, our proposed method is capable of estimating the testing accuracy range of a DNN without any knowledge of training and evaluation data.
\par
While empirical evaluations show that the bounds enclose the \emph{real} test accuracy obtained from the test dataset, we are working to tighten the gap between them. We have defined loss functions using  these bounds which is expected to discourage overfilling and improve performance of the trained network. 

\bibliographystyle{IEEEtran}
\bibliography{citationsDNNsEval}
\end{document}